%% file: main.tex
\documentclass{article}

\usepackage{arxiv}

\usepackage[utf8]{inputenc} % allow utf-8 input
\usepackage[T1]{fontenc}    % use 8-bit T1 fonts
\usepackage{hyperref}       % hyperlinks
\usepackage{url}            % simple URL typesetting
\usepackage{booktabs}       % professional-quality tables
\usepackage{amsmath}
\usepackage{amsfonts}       % blackboard math symbols
\usepackage{amssymb}
\usepackage{nicefrac}       % compact symbols for 1/2, etc.
\usepackage{microtype}      % microtypography
\usepackage{lipsum}
\usepackage{graphicx}
\usepackage{multirow}
\usepackage{multicol}
\usepackage{adjustbox}
\usepackage[table]{xcolor}
\usepackage{tabularx}
\usepackage{cleveref}
\usepackage{caption}

\title{EGM: Efficiently Learning General Motion Tracking Policy for High Dynamic Humanoid Whole-Body Control}

\author{
  Chao Yang\textsuperscript{1,2,*}, Yingkai Sun\textsuperscript{3,*}, Peng Ye\textsuperscript{4}, Xin Chen\textsuperscript{5}, Chong Yu\textsuperscript{2}, Tao Chen\textsuperscript{1,2,$\dagger$} \\[0.1in]
  \normalfont \parbox{0.9\textwidth}{\centering \textsuperscript{1}Shanghai Innovation Institute, \textsuperscript{2}College of Future Information Technology, Fudan University, \textsuperscript{3}College of Intelligent Robotics and Advanced Manufacturing, Fudan University, \textsuperscript{4}The Chinese University of Hong Kong, \textsuperscript{5}ByteDance. \textsuperscript{*}Equal contribution, \textsuperscript{$\dagger$}Corresponding author}
}

\begin{document}
\maketitle

\begin{center}
  \centering
  \includegraphics[width=0.65\linewidth]{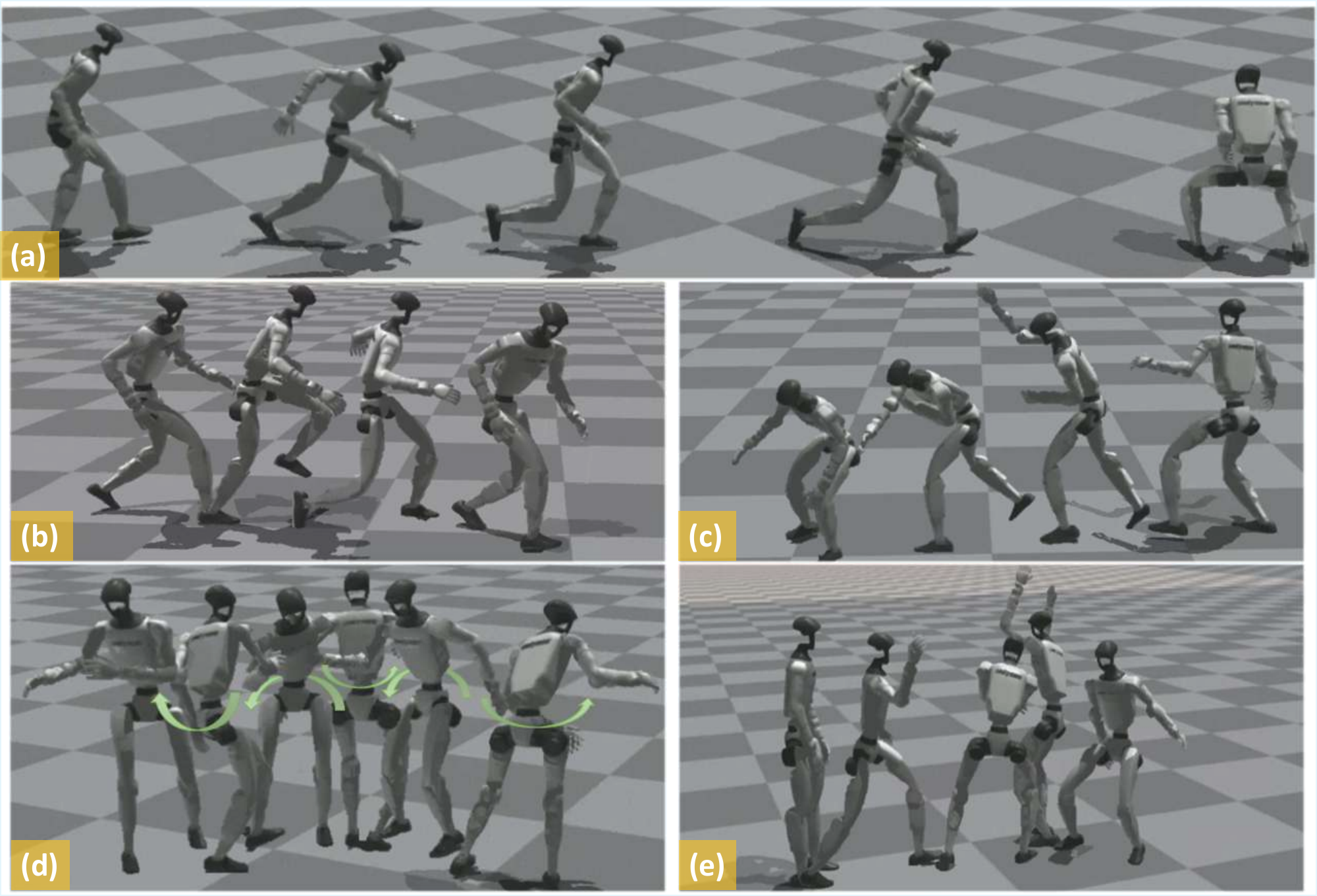}
  \captionof{figure}{We deploy a unified student policy trained with EGM in the simulation environment, achieving high robust stability in tracking highly dynamic motions, including (a) running and sudden stop, (b) hurdling, (c) throwing, (d) quick spinning, and (e) spinning jumps.}
  \label{fig:teaser}
\end{center}

\input{sec/0_abstract}

\input{sec/1_intro}
\input{sec/2_relatedWork}
\input{sec/3_method}

\input{sec/4_experiments}

\input{sec/5_conclusion}

\bibliographystyle{unsrt}
\bibliography{references}

\end{document}

%% file: sec/0_abstract.tex
\begin{abstract}
Learning a general motion tracking policy from human motions shows great potential  for versatile humanoid whole-body control. Conventional approaches are not only inefficient in data utilization and training processes but also exhibit limited performance when tracking highly dynamic motions. To address these challenges, we propose EGM, a framework that enables efficient learning of a general motion tracking policy. EGM integrates four core designs. Firstly, we introduce a Bin-based Cross-motion Curriculum Adaptive Sampling strategy to dynamically orchestrate the sampling probabilities based on tracking error of each motion bin, eficiently balancing the training process across motions with varying dificulty and durations. The sampled data is then processed by our proposed Composite Decoupled Mixture-of-Experts (CDMoE) architecture, which efficiently enhances the ability to track motions from different distributions by grouping experts separately for upper and lower body and decoupling orthogonal experts from shared experts to separately handle dedicated features and general features. Central to our approach is a key insight we identified: for training a general motion tracking policy, data quality and diversity are paramount. Building on these designs, we develop a three-stage curriculum training flow to progressively enhance the policy's robustness against disturbances. Despite training on only 4.08 hours of data, EGM generalized robustly across 49.25 hours of test motions, outperforming baselines on both routine and highly 
dynamic tasks.
\end{abstract}

%% file: sec/1_intro.tex
\section{Introduction}
\label{sec:intro}

Humanoid, by virtue of their human-like morphology, hold significant potential for performing diverse tasks in everyday environments. A fundamental requirement for realizing this is a general whole-body controller capable of executing a wide range of skills. Learning such controller from the vast repertoire of human mocap data presents a compelling pathway.

While significant progress has been made in humanoid motion tracking, most existing work focuses on single motion \cite{peng2018deepmimic, he2025asapaligningsimulationrealworld, xie2025kungfubotphysicsbasedhumanoidwholebody, zhang2025hub, liao2025beyondmimicmotiontrackingversatile}. However, the practical value of humanoid lies in their ability to switch between diverse motions seamlessly, which demands a single policy that adapts to multiple motions. To address this, recent studies have attempted to build universal trackers using architectures like MLPs \cite{he2024omnih2o, ji2024exbody2, ze2025twist, li2025clone, yin2025unitrackerlearninguniversalwholebody}. Some studies has attempted to enhance the capacity of policys by introducing architectures such as MoE and OMoE, or to handle imbalanced data distributions using adaptive sampling strategies \cite{chen2025gmt, han2025kungfubot2}. However, two fundamental issues still remain.

\begin{itemize}
\item \textbf{Redundancy in large-scale Mocap data.} Conventional methods typically utilize dozens of hours of motion data to ensure generalization across various motions, which incurs substantial costs and is inefficient. This inefficiency stems from the inherent imbalance in such datasets (e.g., AMASS \cite{AMASS:ICCV:2019}) as well as the abundance of low-quality or redundant samples. Consequently, the policy fails to prioritize the learning of complex motion features, and its generalization capability remains hindered despite the massive volume of input data.

\item \textbf{Inadequate performance in tracking highly dynamic motions.} Existing methods struggle in this regard for two primary reasons. Firstly, the abundance of redundant or low-quality motions, coupled with inefficient sampling strategies, leads to insufficient learning of these complex patterns. Secondly, while current architectures (e.g., MoE, OMoE \cite{chen2025gmt, han2025kungfubot2}) focus on extracting specialized features for different motions, they often overlook the need for control specialization. Furthermore, these architectures emphasize capturing motion-specific dedicated features but tend to neglect the general features common to all motions, such as balance maintenance. As a result, they fail to achieve precise tracking of highly dynamic motions while ensuring overall stability.
\end{itemize}

To address these problems, we propose EGM, a framework for Efficient General Mimic, which builts upon four core designs. First, we introduce a Bin-based Cross-motion Curriculum Adaptive Sampling strategy, which dynamically adjusts the sampling probability for motion bins, thereby enabling efficient data utilization. Second, we design a Composite Decoupled Mixture-of-Experts (CDMoE) architecture that groups experts separately for the upper and lower body, and decouples orthogonal experts from shared experts to distinctly handle dedicated features and general features. Furthermore, a key insight underlying our approach is that for training a general motion tracking policy, data quality and diversity are paramount. Leveraging this insight, our framework efficiently achieves strong generalization performance using only a small set of high-quality data. Building on these designs, we develop a three-stage curriculum training flow to progressively improve the policy's robustness.

We trained EGM on a small-scale dataset of only 4.08 hours and extensively validated it on a larger test set of 49.25 hours. The experimental results demonstrate that EGM achieves high generalization capability and robust stability in tracking both routine and highly dynamic motions, outperforming baseline methods.To summarize, we make the following contributions:
\begin{itemize}
\item We introduce a Bin-based Cross-motion Curriculum Adaptive Sampling strategy that dynamically balances sample selection across motion types, improving learning efficiency and motion diversity utilization.

\item We design a CDMoE architecture that decouples general and dedicated motion features through body-part expert grouping and expert separation, effectively enhancing the policy's capacity to track motions from varying distributions.

\item We develop a structured three-stage training curriculum that progressively enhances policy robustness and tracking performance across diverse motion types.

\item We demonstrate experimentally that EGM outperforms baseline methods in tracking accuracy and stability, even when trained on only 4.08 hours of data, validating its data efficiency and generalization capability.
\end{itemize}

%% file: sec/2_relatedWork.tex
\section{Related Work}
\label{sec:relatedWork}
%-------------------------------------------------------------------------
\subsection{Humanoid Whole-Body Control}

Traditional model-based approaches have achieved whole-body controllers for robots \cite{miura1984dynamic, sreenath2011compliant, geyer2003positive}. However, they rely on complex system modeling and is typically limited to single tasks. In contrast, learning-based approaches train task-specific whole-body control policies for humanoids through manually designed reward functions, enabling locomotion skills such as walking \cite{radosavovic2024humanoid, radosavovic2024learning, radosavovic2024real, gu2024advancing, chen2024lcp}, jumping \cite{zhang2024wococo, zhuang2024humanoidparkour, xue2025hugwbc}, and fall-recovery \cite{he2025getup, huang2025getup}. Nevertheless, these controllers are often capable of performing only a single motion, requiring a customized reward function and a separately trained policy for each skill. Recent research has developed whole-body controllers capable of handling complex tasks like table tennis \cite{su2025hitterhumanoidtabletennis} and goalkeeping \cite{ren2025humanoidgoalkeeperlearningposition} through methods such as hierarchical planning and learning or by incorporating adversarial motion priors. Yet, the necessity for designing task-specific rewards remains.

In comparison, imitating human motion to design whole-body controllers presents a more efficient approach, as it eliminates the need to engineer specific reward functions for each desired skill.
%-------------------------------------------------------------------------
\subsection{Humanoid Motion Tracking}

DeepMimic \cite{peng2018deepmimic} introduced a framework for learning task-specific controllers from human motion references, thereby reducing the burden of reward engineering. ASAP \cite{he2025asapaligningsimulationrealworld} proposed a real-to-sim pipeline that learns a delta-action model to mitigate sim-to-real gap, although this method involves a complex training process. KungfuBot \cite{xie2025kungfubotphysicsbasedhumanoidwholebody} and HuB \cite{zhang2025hub} focus on enabling humanoid to perform challenging tasks, achieving precise tracking of highly dynamic motions through carefully designed tracking frameworks. However, these approaches are generally limited to imitating individual motions.

In the field of physics-based animation, PHC \cite{luo2023perpetual} demonstrated general motion tracking for virtual characters, but implementing such controllers on real robots is considerably more difficult. OmniH2O \cite{he2024omnih2o}, ExBody2 \cite{ji2024exbody2}, and HumanPlus \cite{fu2024humanplus} successfully implemented general whole-body controllers on humanoid robots, yet their performance in terms of accuracy and stability remains suboptimal. TWIST \cite{ze2025twist} and CLONE \cite{li2025clone} achieved more stable motion tracking, but they are primarily tailored for static tasks in teleoperation scenarios. GMT \cite{chen2025gmt}, UniTracker \cite{yin2025unitrackerlearninguniversalwholebody}, and KungfuBot2 \cite{han2025kungfubot2} enabled tracking of dynamic motions, such as dancing, through methods like adaptive sampling and improved model architectures; however, they struggle to maintain tracking accuracy and stability during highly dynamic maneuvers. BeyondMimic \cite{liao2025beyondmimicmotiontrackingversatile} realized high-precision tracking of individual highly dynamic motions through well-designed tracking objectives and control strategies, and employed a distilled unified diffusion policy for task-specific control, though it requires separate training for each task. 

Building on these works, we aim to propose a method that can efficiently learn a general motion tracking policy capable of accurately and stably tracking highly dynamic motions.

%% file: sec/3_method.tex
\section{Method}

In this section, we present EGM, a framework for Efficient General Mimic. EGM employs a three-stage teacher-student training framework. It first trains a privileged teacher policy using the PPO algorithm \cite{ppo} in two initial stages, followed by distilling a deployable student policy by imitating the teacher via the DAgger algorithm \cite{ross2011reduction}.

We begin by introducing the problem formulation in \cref{sec:problem_formulation}. Then, we present our core components: the Bin-based Cross-motion Curriculum Adaptive Sampling in \cref{sec:bin_sampling} and the Composite Decoupled Mixture-of-Experts in \cref{sec:moe}. Furthermore, we detail the Data Curation process and provide related insights in \cref{sec:data_curation}. Finally, we describe the overall Training Procedure in \cref{sec:training_procedure}.

\begin{figure*}[t]
    \centering
    \includegraphics[width=\textwidth]{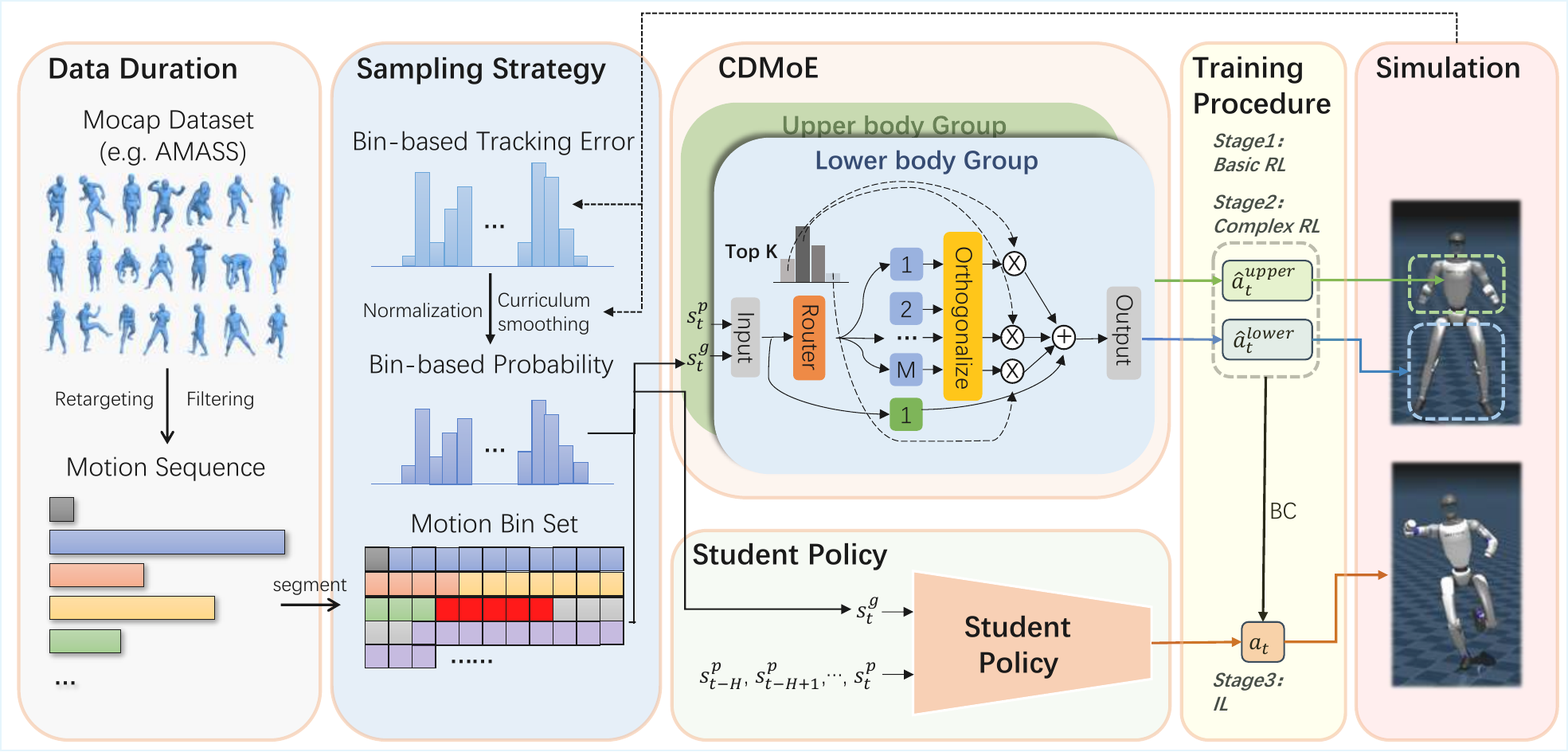}
    \caption{Overview of the EGM framework. First, large-scale Mocap datasets are retargeted to Humanoid, then a small dataset containing diverse and high-quality data is screened out. Next, training is conducted in the simulation environment, where a teacher policy based on the CDMoE architecture is trained using two-stage curriculum reinforcement learning, and then in Stage 3, the teacher policy is distilled into a deployable student policy. Additionally, the Bin-based Cross-motion Curriculum Adaptive Sampling Strategy is used during the training process to achieve dynamic sampling, improving training efficiency.}
    \label{fig:pipeline}
\end{figure*}

\subsection{Problem Formulation}
\label{sec:problem_formulation}

In this work, we employ the 29-degree-of-freedom (DoF) Unitree G1 robot \cite{unitree-g1}. Specifically, we fix the 3 DoFs of each wrist, controlling a total of 23 DoFs. We formulate the problem as a goal-conditioned RL task. We consider a MDP defined by the tuple $\mathcal{M} = \langle S, A, T, R, \gamma \rangle$, where $S$ is the state space, $A$ is the action space, $T$ is the state transition function, $R$ is the reward function, and $\gamma \in [0, 1]$ is the discount factor. The state $s_t \in S$ comprises the proprioceptive information of the robot $s_t^p$ and the goal state $s_t^g$. The action $a_t \in A \subseteq \mathbb{R}^{23}$ represents the target joint positions, which are converted into motor torques via a PD controller to actuate the robot. The transition dynamics $p(s_{t+1}|s_t, a_t)$ governs the evolution of the state. At each time step $t$, the agent receives a dense reward $r_t = R(s_t^p, s_t^g, a_t)$ that evaluates motion tracking performance and provides regularization. The policy $\pi(a_t|s_t)$ aims to maximize the expected discounted return $J = \mathbb{E}\left[ \sum_{t=0}^{T-1} \gamma^t r_t \right]$.

\subsection{Sampling Strategy}
\label{sec:bin_sampling}

Large-scale Mocap datasets like AMASS \cite{AMASS:ICCV:2019} suffer from a imbalanced data distribution, which is reflected not only in the varying quantities of different motion types but also in the differing action categories present within different segments of the same motion sequence. Furthermore, the duration of these motion sequences varies significantly, ranging from a few seconds to several hundred seconds. Previous sampling methods \cite{chen2025gmt,luo2023perpetual} mitigate the impact of imbalanced data distribution to some extent by determining sampling probability based on tracking performance. However, they still sample entire motion clips as units, treating short and long sequences equally. This leads to overfitting on short sequences and insufficient learning of long sequences. Moreover, assigning fixed sampling weights to motions of varying difficulty throughout the entire training process is an inefficient practice.

To address this, we propose a Bin-based Cross-motion Curriculum Adaptive Sampling (BCCAS) strategy, which consists of the following three core designs:

\begin{itemize}
    \item \textbf{Bin-based Cross-motion Global Sampling:} We segment all motion sequences into multiple 1-second motion bins and assign a global ID to each bin. During training, at regular intervals, we perform global sampling across all motion bins. After a bin is sampled, we use it as the starting frame, apply a forward random offset of 0 to 5 seconds, and then extract a subsequent motion clip of up to 20 seconds as the tracking target. This approach flexibly addresses both the issue of varying action types within different segments of the same data and the problem of significantly different sequence lengths across data.
    
    \item \textbf{Sampling Probability based on Composite Tracking Error:} We design a composite tracking error derived from a weighted combination of various error metrics (including position, velocity, etc.). During training, we use an Exponential Moving Average (EMA) to update the tracking error for each bin. After global normalization, these errors are converted into sampling probabilities for each bin.
    
    \item \textbf{Sampling Curriculum:} When calculating the final sampling probability, we introduce a uniform sampling ratio and apply a power function to smooth the probability distribution. During training, by dynamically adjusting the uniform sampling ratio and the smoothing temperature, the policy emphasizes exploration in the early stages. As training progresses, the sampling distribution becomes sharper, gradually focusing more attention on high-error regions. This achieves a balance between exploration and exploitation.
\end{itemize}

Through the BCCAS strategy, we can make the training process more efficient.

\subsection{Composite Decoupled Mixture-of-Experts}
\label{sec:moe}

\begin{figure}[t]
    \centering
    \includegraphics[width=\linewidth]{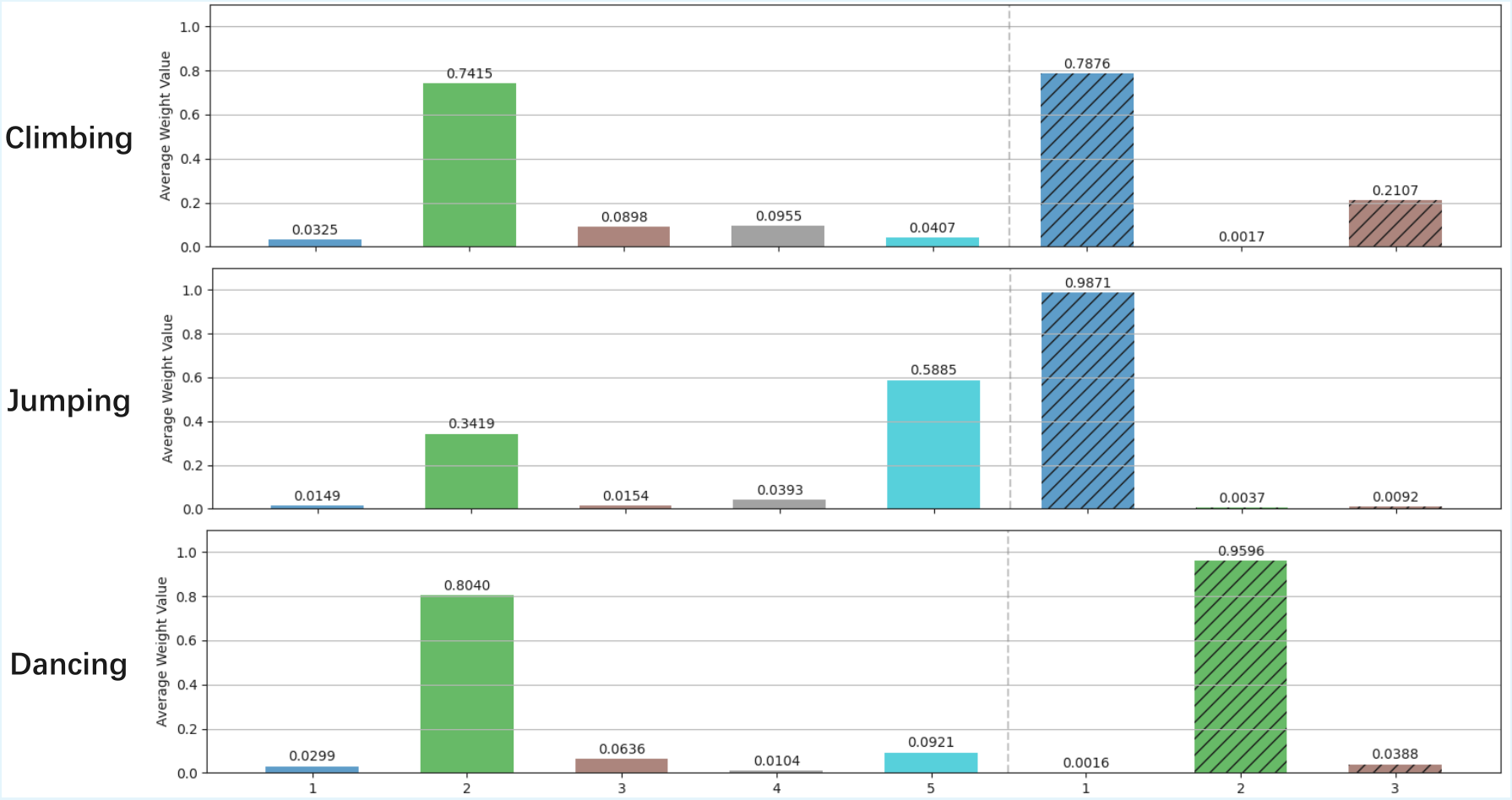}
    \caption{Expert weights distribution across different motion types. Slash represents upperbody experts. Showing how our CDMoE architecture assigns different weights to experts for various motion categories.}
    \label{fig:experts_weight}
\end{figure}

To obtain a general motion tracking policy, the model needs to capture diverse motion features from a large amount of different types of human motion data, including upper and lower body, dynamic and static, explosive and sustained movements, etc., which requires high expressive capability. Previous work has explored using architectures such as Mixture-of-Experts (MoE) \cite{chen2025gmt} and Orthogonal Mixture-of-Experts (OMoE) \cite{han2025kungfubot2} to represent policy. The core idea is to have different experts handle different types of motions to enhance the model's ability to express diverse motion features.
 
However, for humanoid, the motion features of the upper and lower body are often different, and the control difficulty of the lower body is usually higher than that of the upper body. Using the same experts to process both upper and lower body actions can disperse the model's attention and reduce its expressive capability. Additionally, these architectures emphasize specialization, striving to capture dedicated features of specific motions, but often overlook general features common to all movements, such as balance maintenance.
 
To address these issues, we propose the Composite Decoupled Mixture-of-Experts (CDMoE) architecture. Similar to MoE, CDMoE consists of expert networks and a gating network. The expert networks take the robot's proprioceptive information $s_t^p$ and the goal state $s_t^g$ as inputs and output actions $a_t$. The gating network receives the same observations as input and outputs weights for experts. The final model output is a weighted combination of the expert outputs. However, CDMoE includes two core improvements:
 
\begin{itemize}
    \item \textbf{General-Specialized Decoupled Experts:} Referencing the work of Han et al., we constrain the outputs of various experts to be mutually orthogonal through the Gram-Schmidt (GS) process \cite{leon2013gram, hendawy2024multitaskreinforcementlearningmixture}, resulting in a set of $M$ orthogonal experts $\{E_i^o\}_{i=1}^{M}$ for extracting more specialized motion features. On this basis, we introduce a top-$k$ mechanism: after the gating network outputs expert weights $\{w_i\}_{i=1}^{M}$, we only select the $k$ orthogonal experts with the highest weights $\{E_i^o \mid i \in \mathcal{I}_{\text{top-}k}\}$, where $\mathcal{I}_{\text{top-}k} = \arg\max_{\substack{\mathcal{I} \subseteq \{1,2,\dots,M\} \\ |\mathcal{I}|=k}} \sum_{i \in \mathcal{I}} w_i$, and normalize the weights corresponding to these $k$ experts, i.e., $\hat{w}_i = \frac{w_i}{\sum_{j \in \mathcal{I}_{\text{top-}k}} w_j}$. The orthogonal motion features are finally obtained through weighted combination:

   \begin{equation}
P_{\text{orthogonal}} = \sum_{i \in \mathcal{I}_{\text{top-}k}} \hat{w}_i \cdot E_i^o
\label{eq:orthogonal}
\end{equation} 
   
   Furthermore, we introduce a shared expert $E^s$. The shared expert directly outputs shared motion features $P_{\text{shared}}$ without going through the GS process and the gating network. The combined motion features are obtained by combining the orthogonal motion features with the shared motion features:
   
   \begin{equation}
P_{\text{comb}} = P_{\text{orthogonal}} + P_{\text{shared}}
\label{eq:comb}
\end{equation} 
   
   Finally, the action output is obtained through a unified output network:
   
   \begin{equation}
a_t = f_{\text{out}}(P_{\text{comb}})
\label{eq:action}
\end{equation}
   
    \item \textbf{Upper and Lower Body Grouped Experts:} Our control target is a 23-DoF humanoid, with 8 DoFs in the upper body and 15 DoFs in the lower body. Considering the different dynamic characteristics of the upper and lower body, we introduce two independent groups of experts to control the upper and lower body respectively, promoting specialization in control. Each group of experts includes their own orthogonal experts, shared experts, and routing networks, and ultimately only outputs actions for the corresponding joints, but they all receive the same whole-body observations. Additionally, since the lower body is more difficult to control than the upper body and has more joints, we allocate more experts to the lower body.
\end{itemize}
   
As can be seen from \cref{fig:experts_weight}, CDMoE assigns different weights to experts when processing different types of motions. Furthermore, these weight distributions are often quite sharp, indicating a high level of specialization among the experts.
   
Through the design of CDMoE, the policy can simultaneously extract dedicated motion features and general motion features, enhance the level of control specialization, thereby improve the model's capacity to handle motions from different distributions. Moreover, the upper and lower body grouped experts allow us to better allocate computational resources, achieving training efficiency.

\subsection{Data Curation}
\label{sec:data_curation}

We collected human motion data from AMASS \cite{AMASS:ICCV:2019} and LAFAN1 \cite{harvey2020lafan}, and retargeted the motions into robot motions following the shape-and-motion two-stage retargeting process \cite{he2024h2o}. However, the original retargeted dataset contains numerous unachievable motions (such as climbing stairs and sitting on chairs) and motions with retargeting failures (such as floating and jittering), thus requiring dataset filtering. We organized the training dataset in two distinct ways: 

\begin{enumerate}
    \item Manual screening to obtain a small yet high-quality dataset $\mathcal{D}_{\text{small}}$ (containing 674 diverse motions with a total duration of 4.08 hours). 
    \item Initial filtering of the entire dataset based on rules (e.g., keywords, height). Subsequently, we first trained a policy on the $\mathcal{D}_{\text{small}}$ dataset and used this policy to evaluate the initially filtered dataset. Based on the evaluation results, we further removed motions with excessively high tracking errors, yielding a larger training dataset $\mathcal{D}_{\text{large}}$ (containing 10,393 motions with a total duration of 27.47 hours). 
\end{enumerate}

We trained and evaluated the policies using the two datasets separately, and ultimately found that the policy trained on $\mathcal{D}_{\text{small}}$ outperforms the one trained on $\mathcal{D}_{\text{large}}$ in both generalization and accuracy. 

This is because large-scale Mocap datasets like AMASS \cite{AMASS:ICCV:2019} contain substantial low-quality or redundant data---for instance, dozens of identical walking motions. Additionally, failed retargeting leads to some low-quality motions. As a result, even though datasets may include tens of thousands of motions, the actual effective data might be limited. Meanwhile, the massive amount of low-quality or redundant data diverts the policy's attention during training, resulting in inefficient training. In contrast, although the manually screened dataset $\mathcal{D}_{\text{small}}$ is modest in scale, it features low redundancy, high diversity, and superior quality, thereby achieving better training outcomes. 

Therefore, we chose to use the small yet high-quality dataset $\mathcal{D}_{\text{small}}$ for training, and put forward a key insight: for training a general motion tracking policy, data quality and diversity are paramount.

\subsection{Training Procedure}
\label{sec:training_procedure}

We adopt a teacher-student framework similar to previous works for training \cite{ji2024exbody2,he2024omnih2o,chen2025gmt,han2025kungfubot2}. However, unlike the prior two-stage framework, we implement a three-stage training process, with the key difference lying in the two-stage training of the teacher policy. 

We first train the teacher policy in Stage 1 using the PPO algorithm \cite{ppo}, then load the trained checkpoint for Stage 2 training. The main distinction between the two stages is that Stage 2 introduces domain randomization \cite{Peng_2018} and complex terrains (including rough and uneven terrains, etc.). Additionally, in terms of the reward function, Stage 2 incorporates more refined tracking terms and stricter penalty terms compared to Stage 1. This design aims to realize a curriculum training flow: the policy first learns basic motion tracking capabilities in simple environments, and then enhances the precision, smoothness, and robustness of motion tracking through more stringent reward functions, environmental disturbances, and terrain challenges. 

The teacher policy requires certain privileged observations that are difficult to obtain in the real world, making it unsuitable for direct deployment. Therefore, after completing the training of the teacher policy, we use the DAgger algorithm \cite{ross2011reduction} to distill the teacher policy into a deployable student policy. Unlike the teacher policy, the student policy's input does not include privileged observations. To compensate for the resulting information loss, we design a history encoder \cite{he2024omnih2o} for the student policy, which encodes a certain number of time steps of proprioceptive history as input.

%% file: sec/4_experiments.tex
\section{Experiments}

\label{sec:experiments}

In this section, we conduct comprehensive experiments to validate the effectiveness of our proposed EGM framework. We first introduce the experimental setups in \cref{sec:exp_setup}, then compare our method with baseline approache in \cref{sec:baseline}. Subsequently, we perform detailed ablation studies to analyze the contribution of each component in \cref{sec:ablation}.
% Insert the ablation baseline table
\input{table/ablation_baseline}

\subsection{Experimental Setups}
\label{sec:exp_setup}

We use the retargeted and filtered AMASS \cite{AMASS:ICCV:2019} and LAFAN1 \cite{harvey2020lafan} datasets, specifically the $
\mathcal{D}_{\text{small}}$ subset, to train policies for the 23-DoF Unitree G1 robot \cite{unitree-g1} in the IsaacGym \cite{isaac} simulator. The trained policies are then evaluated on both the $
\mathcal{D}_{\text{eval}}$ dataset (excluding training data, containing 13,596 motions with a total duration of 49.25 hours) and the $
\mathcal{D}_{\text{small}}$ training set.

The evaluation metrics include:
\begin{itemize}
    \item $E_{\text{mpkpe}}$: Mean Per Keypoint Position Error, measured in mm
    \item $E_{\text{umpjpe}}$: Upper body Mean Per Joint Position Error, measured in rad
    \item $E_{\text{lmpjpe}}$: Lower body Mean Per Joint Position Error, measured in rad
    \item $E_{\text{acc\_dist}}$: Average joint acceleration error, measured in mm/frame$^2$
    \item $E_{\text{vel\_dist}}$: Average joint velocity error, measured in mm/frame
\end{itemize}

\subsection{Baseline}
\label{sec:baseline}

We compare our EGM framework with GMT \cite{chen2025gmt}. We re-implemented GMT and trained it on the same $\mathcal{D}_{\text{small}}$ dataset, then evaluated it in the same simulation environment. As shown in \cref{tab:ablation_baseline}, EGM outperforms GMT across all evaluation metrics.

\subsection{Ablation Studies}
\label{sec:ablation}

\paragraph{Sampling Strategy.}
\label{sec:ablation_sampling}

\begin{figure}[t]
    \centering
    \includegraphics[width=0.5\linewidth]{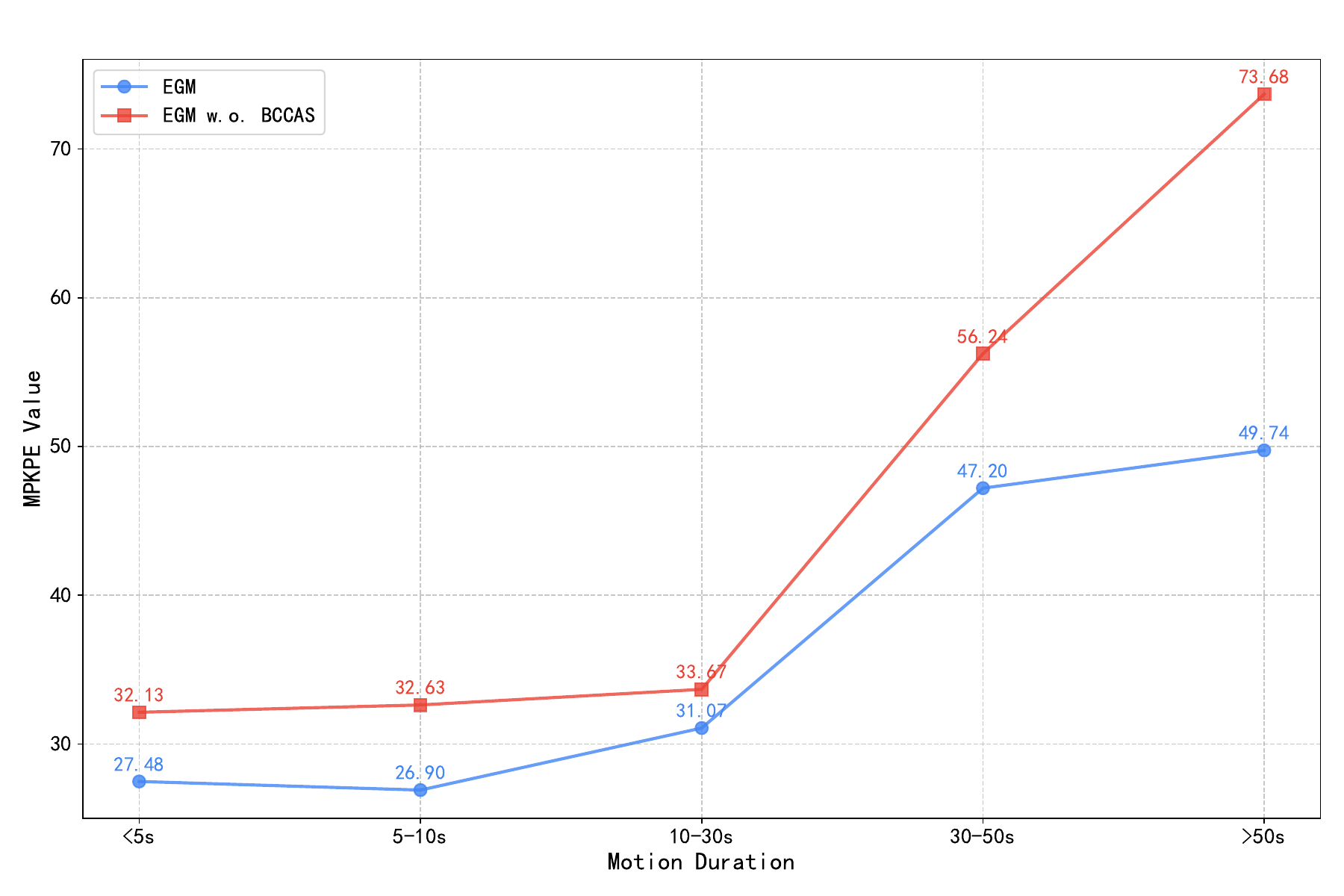}
    \caption{$E_{\text{mpkpe}}$ across motion durations. After using BCCAS, the policy has obvious advantages in tracking long-sequence motions.}
    \label{fig:sampling_strategy}
\end{figure}

\begin{figure*}[t]
    \centering
    \includegraphics[width=\textwidth]{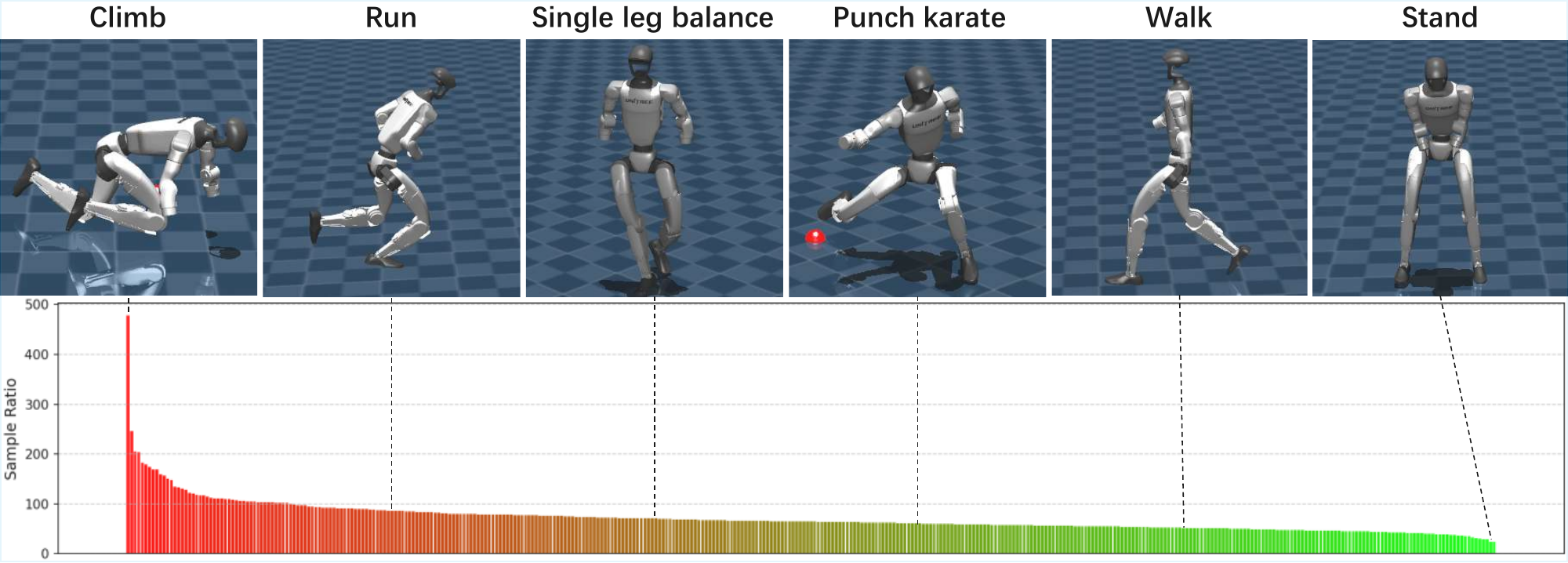}
    \caption{The distribution of Sample ratio in the training data, as well as the typical motions with different Sample ratios. Motions with a larger Sample ratio tend to be more difficult, indicating that BCCAS has successfully achieved a higher concentration of attention on more difficult movements.}
    \label{fig:sample_ratio}
\end{figure*}

As shown in \cref{tab:ablation_baseline}(a), after using BCCAS, the policy shows improvements across all metrics, effectively enhancing the tracking performance. Specifically, we conducted ablation studies on BCCAS across motions with different durations. As illustrated in \cref{fig:sampling_strategy}, compared to the policy without BCCAS, the policy using BCCAS not only achieves lower tracking errors across all motion durations but also demonstrates more significant advantages in longer-duration motions. This validates the role of BCCAS in addressing insufficient training for long-sequence motions.

Furthermore, to verify whether BCCAS truly provides more attention to more difficult and rarer motions, we statistically analyzed the number of times each motion data's bins were sampled during training. By dividing the sampling count by the duration of the motion data, we obtain a sample ratio. Ideally, after using BCCAS, the sample ratio should represent the average difficulty of a motion data. As demonstrated in \cref{fig:sample_ratio}, motions with higher sample ratios tend to be more difficult (such as running, hurdling, etc.), while motions with lower sample ratios are less challenging (such as walking, standing, etc.). Additionally, in our statistics, the motion data with the highest sample ratio is a climbing sequence, which is exactly the most scarce type of contact-rich ground motion, and BCCAS provides it with significantly more attention relative to other motions. These results indicate that BCCAS successfully provides more attention to more difficult and rarer motions, achieving more efficient learning.

\paragraph{Composite Decoupled Mixture-of-Experts.}
\label{sec:ablation_moe}

\begin{figure}[t]
    \centering
    \includegraphics[width=\linewidth]{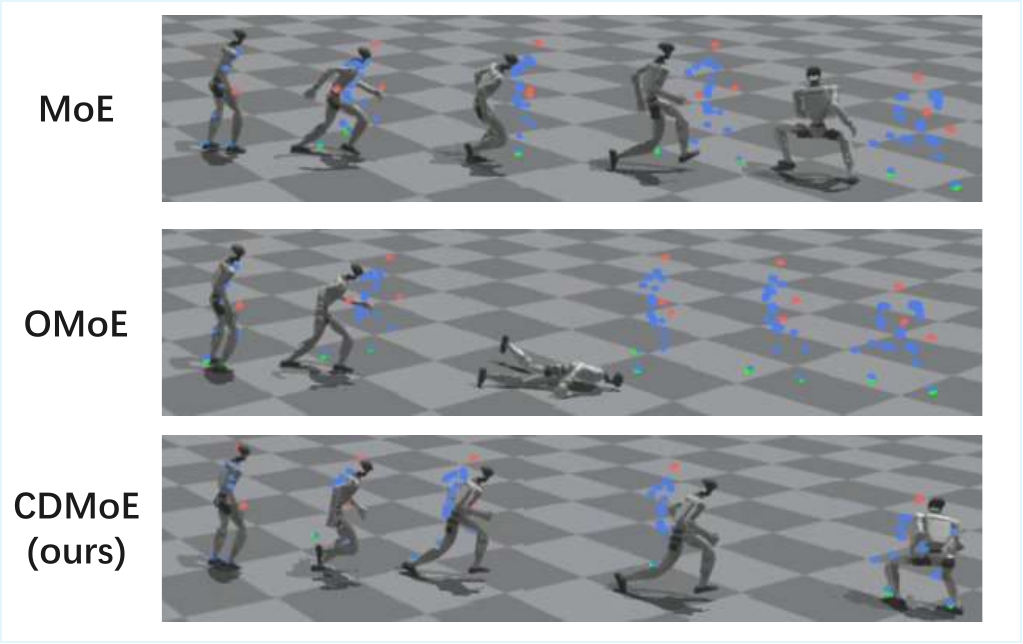}
    \caption{Performance of different policy architectures when tracking a running-to-sudden-stop motion. CDMoE achieves the best performance in both stability and accuracy.}
    \label{fig:cdmoe_com}
\end{figure}

We analyze the impact of our CDMoE architecture by comparing it with MoE \cite{chen2025gmt} and OMoE \cite{han2025kungfubot2} architectures. As shown in \cref{tab:ablation_baseline}(b), the policy using CDMoE outperforms those using MoE or OMoE across all metrics, particularly in velocity and acceleration tracking errors. Furthermore, to intuitively investigate the role of CDMoE, we use policies with MoE, OMoE, and CDMoE respectively to track some highly dynamic motions and compare their performance. As illustrated in \cref{fig:cdmoe_com}, when tracking the same running-to-sudden-stop motion, the MoE policy, although not falling, exhibits conservative behavior and fails to track the faster running speed, resulting in significant tracking errors. The OMoE policy accelerates to improve motion performance but cannot maintain stability and directly falls. In contrast, the CDMoE policy accurately tracks the entire motion sequence while maintaining both high speed and stability. These results indicate that CDMoE can enhance the policy's capacity, enabling it to ensure both stability and precision when tracking highly dynamic motions.

\paragraph{Training Dataset.}
\label{sec:ablation_dataset}

To validate the hypothesis that ``data quality and diversity are paramount,'' we conducted ablation studies by training policies under three different conditions: (1) using $\mathcal{D}_{\text{large}}$, a dataset containing extensive motion data but only filtered through automated processes; (2) using $\mathcal{D}_{\text{small}}$, a manually curated dataset with diverse and high-quality motion data; and (3) pre-training on $\mathcal{D}_{\text{large}}$ followed by fine-tuning on $\mathcal{D}_{\text{small}}$. For fair evaluation, we used the ASAP-provided motion dataset $\mathcal{D}_{\text{asap}}$ \cite{he2025asapaligningsimulationrealworld} for assessment. As shown in \cref{tab:data_ablation}, both policies trained with $\mathcal{D}_{\text{large}}$ demonstrate inferior tracking capabilities compared to the policy trained with $\mathcal{D}_{\text{small}}$, despite incurring higher data processing, storage, and computational costs. This indicates that for training a general motion tracking policy, data quality and diversity are crucial factors, not merely the quantity of data.

\paragraph{Training Procedure.}
\label{sec:ablation_training}
\input{table/data_ablation}
\input{table/training_ablation}
To validate the effectiveness of our three-stage curriculum training flow, we compared it with a two-stage approach that directly introduces fine-grained tracking rewards, domain randomization, and complex terrains. Both methods were trained for the same time and with identical data to train teacher policies, followed by evaluation. As shown in \cref{tab:ablation_baseline}(c), the teacher policy trained with our three-stage approach demonstrates significantly superior tracking capabilities compared to the policy trained with the two-stage method. This is because the two-stage approach immediately demands high-precision and high-robustness tracking from the policy when it has not yet acquired basic tracking abilities, resulting in the policy being unable to consolidate fundamental motion tracking skills while struggling to learn more refined and robust tracking techniques. In contrast, the three-stage curriculum training flow enables the policy to first learn basic motion tracking capabilities in simple environments, then progressively acquire more refined and robust tracking skills, achieving better training outcomes.
Furthermore, we evaluated the three-stage policies trained using EGM. As shown in \cref{tab:training_ablation}, the teacher policy's tracking capabilities improved significantly after Stage2 compared to Stage1, confirming the effectiveness of the curriculum training flow. And the student policy in Stage3 still retains most of the performance of the teacher policy, although it loses some privileged observations.

%% file: table/ablation_baseline.tex
\begin{table*}[htbp]
\caption{The evaluation results on the test dateset and train dateset, where all policys are trained using the same $\mathcal{D}_{\text{small}}$ dataset. In all baseline comparisons and ablation studies, considering that the student policy imitates the teacher policy using the exact same distillation method, we chose to evaluate the teacher policy.}
\label{tab:ablation_baseline}
\centering
\renewcommand{\arraystretch}{0.9}
\resizebox{0.95\textwidth}{!}{
\begin{tabular}{lcccccccccc}
\toprule
\multicolumn{1}{l}{} & \multicolumn{5}{c}{\cellcolor{blue!10}\textbf{Test Dataset}} & \multicolumn{5}{c}{\cellcolor{green!10}\textbf{Train Dataset}} \\
\cmidrule(r){1-1}\cmidrule(r){2-6}\cmidrule(r){7-11}
\textbf{Method} & $E_{\text{mpkpe}}$ $\downarrow$ & $E_{\text{umpjpe}}$ $\downarrow$ & $E_{\text{lmpjpe}}$ $\downarrow$ & $E_{\text{acc\_dist}}$ $\downarrow$ & $E_{\text{vel\_dist}}$ $\downarrow$ & $E_{\text{mpkpe}}$ $\downarrow$ & $E_{\text{umpjpe}}$ $\downarrow$ & $E_{\text{lmpjpe}}$ $\downarrow$ & $E_{\text{acc\_dist}}$ $\downarrow$ & $E_{\text{vel\_dist}}$ $\downarrow$ \\
\cmidrule(r){1-1}\cmidrule(r){2-6}\cmidrule(r){7-11}
\rowcolor{lightgray}
\multicolumn{11}{l}{\textbf{Baseline}} \\
\cmidrule(r){1-1}\cmidrule(r){2-6}\cmidrule(r){7-11}
GMT & $67.09_{\scriptstyle\pm 0.11}$ & $0.07197_{\scriptstyle\pm 0.00012}$ & $0.06373_{\scriptstyle\pm 0.00011}$ & $19.78_{\scriptstyle\pm 0.14}$ & $14.21_{\scriptstyle\pm 0.05}$ & $53.13_{\scriptstyle\pm 0.26}$ & $0.05825_{\scriptstyle\pm 0.00034}$ & $0.04962_{\scriptstyle\pm 0.00022}$ & $13.06_{\scriptstyle\pm 0.25}$ & $11.08_{\scriptstyle\pm 0.09}$ \\
EGM (ours) & $57.35_{\scriptstyle\pm 0.08}$ & $0.06204_{\scriptstyle\pm 0.00010}$ & $0.05412_{\scriptstyle\pm 0.00008}$ & $18.12_{\scriptstyle\pm 0.06}$ & $12.95_{\scriptstyle\pm 0.02}$ & $45.45_{\scriptstyle\pm 0.23}$ & $0.05046_{\scriptstyle\pm 0.00020}$ & $0.04200_{\scriptstyle\pm 0.00026}$ & $12.12_{\scriptstyle\pm 0.10}$ & $9.99_{\scriptstyle\pm 0.09}$ \\
\cmidrule(r){1-1}\cmidrule(r){2-6}\cmidrule(r){7-11}
\rowcolor{lightgray}
\multicolumn{11}{l}{\textbf{(a) Ablations on BCCAS}} \\
\cmidrule(r){1-1}\cmidrule(r){2-6}\cmidrule(r){7-11}
EGM w.o. BCCAS & $71.04_{\scriptstyle\pm 0.42}$ & $0.07510_{\scriptstyle\pm 0.00044}$ & $0.06826_{\scriptstyle\pm 0.00041}$ & $20.71_{\scriptstyle\pm 0.09}$ & $14.70_{\scriptstyle\pm 0.09}$ & $55.60_{\scriptstyle\pm 0.18}$ & $0.06057_{\scriptstyle\pm 0.00010}$ & $0.05219_{\scriptstyle\pm 0.00025}$ & $12.92_{\scriptstyle\pm 0.21}$ & $10.88_{\scriptstyle\pm 0.03}$ \\
EGM w.o. Sampling Curriculum & $62.18_{\scriptstyle\pm 0.19}$ & $0.06696_{\scriptstyle\pm 0.00019}$ & $0.05890_{\scriptstyle\pm 0.00019}$ & $18.98_{\scriptstyle\pm 0.02}$ & $13.45_{\scriptstyle\pm 0.02}$ & $52.99_{\scriptstyle\pm 0.83}$ & $0.05753_{\scriptstyle\pm 0.00079}$ & $0.04987_{\scriptstyle\pm 0.00086}$ & $12.26_{\scriptstyle\pm 0.18}$ & $10.30_{\scriptstyle\pm 0.09}$ \\
EGM (ours) & $57.35_{\scriptstyle\pm 0.08}$ & $0.06204_{\scriptstyle\pm 0.00010}$ & $0.05412_{\scriptstyle\pm 0.00008}$ & $18.12_{\scriptstyle\pm 0.06}$ & $12.95_{\scriptstyle\pm 0.02}$ & $45.45_{\scriptstyle\pm 0.23}$ & $0.05046_{\scriptstyle\pm 0.00020}$ & $0.04200_{\scriptstyle\pm 0.00026}$ & $12.12_{\scriptstyle\pm 0.10}$ & $9.99_{\scriptstyle\pm 0.09}$ \\
\cmidrule(r){1-1}\cmidrule(r){2-6}\cmidrule(r){7-11}
\rowcolor{lightgray}
\multicolumn{11}{l}{\textbf{(b) Ablations on CDMoE}} \\
\cmidrule(r){1-1}\cmidrule(r){2-6}\cmidrule(r){7-11}
EGM w MoE & $64.74_{\scriptstyle\pm 0.21}$ & $0.06939_{\scriptstyle\pm 0.00021}$ & $0.06155_{\scriptstyle\pm 0.00022}$ & $19.02_{\scriptstyle\pm 0.13}$ & $13.73_{\scriptstyle\pm 0.07}$ & $46.66_{\scriptstyle\pm 0.20}$ & $0.05198_{\scriptstyle\pm 0.00020}$ & $0.04299_{\scriptstyle\pm 0.00021}$ & $12.71_{\scriptstyle\pm 0.13}$ & $10.51_{\scriptstyle\pm 0.04}$ \\
EGM w OMoE & $60.34_{\scriptstyle\pm 0.08}$ & $0.06523_{\scriptstyle\pm 0.00006}$ & $0.05698_{\scriptstyle\pm 0.00009}$ & $19.73_{\scriptstyle\pm 0.08}$ & $14.01_{\scriptstyle\pm 0.03}$ & $47.64_{\scriptstyle\pm 0.67}$ & $0.05285_{\scriptstyle\pm 0.00064}$ & $0.04407_{\scriptstyle\pm 0.00070}$ & $14.33_{\scriptstyle\pm 0.48}$ & $11.53_{\scriptstyle\pm 0.19}$ \\
EGM (ours) & $57.35_{\scriptstyle\pm 0.08}$ & $0.06204_{\scriptstyle\pm 0.00010}$ & $0.05412_{\scriptstyle\pm 0.00008}$ & $18.12_{\scriptstyle\pm 0.06}$ & $12.95_{\scriptstyle\pm 0.02}$ & $45.45_{\scriptstyle\pm 0.23}$ & $0.05046_{\scriptstyle\pm 0.00020}$ & $0.04200_{\scriptstyle\pm 0.00026}$ & $12.12_{\scriptstyle\pm 0.10}$ & $9.99_{\scriptstyle\pm 0.09}$ \\
\cmidrule(r){1-1}\cmidrule(r){2-6}\cmidrule(r){7-11}
\rowcolor{lightgray}
\multicolumn{11}{l}{\textbf{(c) Ablations on Training Procedure}} \\
\cmidrule(r){1-1}\cmidrule(r){2-6}\cmidrule(r){7-11}
EGM w.o. Three-Stage-Training & $110.42_{\scriptstyle\pm 0.10}$ & $0.11985_{\scriptstyle\pm 0.00012}$ & $0.10394_{\scriptstyle\pm 0.00009}$ & $24.97_{\scriptstyle\pm 0.13}$ & $18.40_{\scriptstyle\pm 0.03}$ & $83.61_{\scriptstyle\pm 0.54}$ & $0.09241_{\scriptstyle\pm 0.00069}$ & $0.07758_{\scriptstyle\pm 0.00045}$ & $13.28_{\scriptstyle\pm 0.30}$ & $12.60_{\scriptstyle\pm 0.10}$ \\
EGM (ours) & $57.35_{\scriptstyle\pm 0.08}$ & $0.06204_{\scriptstyle\pm 0.00010}$ & $0.05412_{\scriptstyle\pm 0.00008}$ & $18.12_{\scriptstyle\pm 0.06}$ & $12.95_{\scriptstyle\pm 0.02}$ & $45.45_{\scriptstyle\pm 0.23}$ & $0.05046_{\scriptstyle\pm 0.00020}$ & $0.04200_{\scriptstyle\pm 0.00026}$ & $12.12_{\scriptstyle\pm 0.10}$ & $9.99_{\scriptstyle\pm 0.09}$ \\
\bottomrule
\end{tabular}}

\end{table*}

%% file: table/data_ablation.tex
\begin{table}[htbp]
\caption{Evaluation results on $\mathcal{D}_{\text{asap}}$ using different training datasets.}
\label{tab:data_ablation}
\centering
\renewcommand{\arraystretch}{1.1}
\begin{tabular}{lccc}
\toprule
\textbf{Train Dataset} & $E_{\text{mpkpe}}$ & $E_{\text{acc\_dist}}$ & $E_{\text{vel\_dist}}$ \\
\midrule
$\mathcal{D}_{\text{large}}$ & $33.58_{\scriptstyle\pm 1.86}$ & $13.11_{\scriptstyle\pm 0.84}$ & $8.76_{\scriptstyle\pm 0.41}$ \\
$\mathcal{D}_{\text{large}} \rightarrow \mathcal{D}_{\text{small}}$ & $31.75_{\scriptstyle\pm 0.44}$ & $14.10_{\scriptstyle\pm 0.89}$ & $9.28_{\scriptstyle\pm 0.26}$ \\
$\mathcal{D}_{\text{small}}$ (ours) & $31.07_{\scriptstyle\pm 0.27}$ & $11.23_{\scriptstyle\pm 0.18}$ & $7.80_{\scriptstyle\pm 0.25}$ \\
\bottomrule
\end{tabular}
\end{table}

%% file: table/training_ablation.tex
\begin{table}[htbp]
\caption{Evaluation results of EGM policies at different training stages on $\mathcal{D}_{\text{eval}}$.}
\label{tab:training_ablation}
\centering
\renewcommand{\arraystretch}{1.1}
\setlength{\tabcolsep}{2pt}
\begin{tabular}{lccc}
\toprule
\textbf{Train Procedure} & $E_{\text{mpkpe}}$ & $E_{\text{acc\_dist}}$ & $E_{\text{vel\_dist}}$ \\
\midrule
Stage1 & $78.06_{\scriptstyle\pm 0.15}$ & $19.20_{\scriptstyle\pm 0.12}$ & $14.18_{\scriptstyle\pm 0.03}$ \\
Stage1+2 & $57.35_{\scriptstyle\pm 0.08}$ & $18.12_{\scriptstyle\pm 0.06}$ & $12.95_{\scriptstyle\pm 0.02}$ \\
Stage1+2+3 & $65.15_{\scriptstyle\pm 0.09}$ & $19.83_{\scriptstyle\pm 0.11}$ & $14.02_{\scriptstyle\pm 0.05}$ \\
\bottomrule
\end{tabular}
\end{table}

%% file: sec/5_conclusion.tex
\section{Conclusion}

In this work, we proposed EGM, a framework for Efficient General Mimic that enables effective learning of general motion tracking policy for humanoid. Through four core designs—Bin-based Cross-motion Curriculum Adaptive Sampling, Composite Decoupled Mixture-of-Experts architecture, emphasis on data quality and diversity, and a three-stage curriculum training flow—EGM addresses the fundamental challenges of data redundancy and inadequate performance in tracking highly dynamic motions. Our experimental results demonstrate that despite training on only 4.08 hours of carefully curated data, EGM achieves robust generalization across 49.25 hours of diverse test motions, outperforming baseline methods on both routine and highly dynamic tasks.

%% file: references.bib
@String(TOG= {ACM Trans. Graph.})

@String(TOG   = {ACM TOG})

@article{peng2018deepmimic,
  title={Deepmimic: Example-guided deep reinforcement learning of physics-based character skills},
  author={Peng, Xue Bin and Abbeel, Pieter and Levine, Sergey and Van de Panne, Michiel},
  journal={ACM Transactions On Graphics (TOG)},
  volume={37},
  number={4},
  pages={1--14},
  year={2018},
  publisher={ACM New York, NY, USA}
}

@misc{he2025asapaligningsimulationrealworld,
      title={ASAP: Aligning Simulation and Real-World Physics for Learning Agile Humanoid Whole-Body Skills}, 
      author={Tairan He and Jiawei Gao and Wenli Xiao and Yuanhang Zhang and Zi Wang and Jiashun Wang and Zhengyi Luo and Guanqi He and Nikhil Sobanbab and Chaoyi Pan and Zeji Yi and Guannan Qu and Kris Kitani and Jessica Hodgins and Linxi "Jim" Fan and Yuke Zhu and Changliu Liu and Guanya Shi},
      year={2025},
      eprint={2502.01143},
      archivePrefix={arXiv},
      primaryClass={cs.RO},
      url={https://arxiv.org/abs/2502.01143}, 
}

@misc{xie2025kungfubotphysicsbasedhumanoidwholebody,
      title={KungfuBot: Physics-Based Humanoid Whole-Body Control for Learning Highly-Dynamic Skills}, 
      author={Weiji Xie and Jinrui Han and Jiakun Zheng and Huanyu Li and Xinzhe Liu and Jiyuan Shi and Weinan Zhang and Chenjia Bai and Xuelong Li},
      year={2025},
      eprint={2506.12851},
      archivePrefix={arXiv},
      primaryClass={cs.RO},
      url={https://arxiv.org/abs/2506.12851}, 
}

@article{zhang2025hub,
  title={HuB: Learning Extreme Humanoid Balance},
  author={Zhang, Tong and Zheng, Boyuan and Nai, Ruiqian and Hu, Yingdong and Wang, Yen-Jen and Chen, Geng and Lin, Fanqi and Li, Jiongye and Hong, Chuye and Sreenath, Koushil and Gao, Yang},
  journal={arXiv preprint arXiv:2505.07294},
  year={2025}
}

@misc{liao2025beyondmimicmotiontrackingversatile,
      title={BeyondMimic: From Motion Tracking to Versatile Humanoid Control via Guided Diffusion}, 
      author={Qiayuan Liao and Takara E. Truong and Xiaoyu Huang and Guy Tevet and Koushil Sreenath and C. Karen Liu},
      year={2025},
      eprint={2508.08241},
      archivePrefix={arXiv},
      primaryClass={cs.RO},
      url={https://arxiv.org/abs/2508.08241}, 
}

@article{he2024omnih2o,
  title={Omnih2o: Universal and dexterous human-to-humanoid whole-body teleoperation and learning},
  author={He, Tairan and Luo, Zhengyi and He, Xialin and Xiao, Wenli and Zhang, Chong and Zhang, Weinan and Kitani, Kris and Liu, Changliu and Shi, Guanya},
  journal={arXiv preprint arXiv:2406.08858},
  year={2024}
}

@article{ji2024exbody2,
  title={Exbody2: Advanced expressive humanoid whole-body control},
  author={Ji, Mazeyu and Peng, Xuanbin and Liu, Fangchen and Li, Jialong and Yang, Ge and Cheng, Xuxin and Wang, Xiaolong},
  journal={arXiv preprint arXiv:2412.13196},
  year={2024}
}

@article{ze2025twist,
title={TWIST: Teleoperated Whole-Body Imitation System},
author= {Yanjie Ze and Zixuan Chen and João Pedro Araújo and Zi-ang Cao and Xue Bin Peng and Jiajun Wu and C. Karen Liu},
year= {2025},
journal= {arXiv preprint arXiv:2505.02833}
}

@misc{li2025clone,
  title={CLONE: Closed-Loop Whole-Body Humanoid Teleoperation for Long-Horizon Tasks}, 
  author={Yixuan Li and Yutang Lin and Jieming Cui and Tengyu Liu and Wei Liang and Yixin Zhu and Siyuan Huang},
  journal={arXiv preprint arXiv:2506.08931}, 
  year={2025}
}

@misc{yin2025unitrackerlearninguniversalwholebody,
      title={UniTracker: Learning Universal Whole-Body Motion Tracker for Humanoid Robots}, 
      author={Kangning Yin and Weishuai Zeng and Ke Fan and Zirui Wang and Qiang Zhang and Zheng Tian and Jingbo Wang and Jiangmiao Pang and Weinan Zhang},
      year={2025},
      eprint={2507.07356},
      archivePrefix={arXiv},
      primaryClass={cs.RO},
      url={https://arxiv.org/abs/2507.07356}, 
}

@misc{chen2025gmt,
      title={GMT: General Motion Tracking for Humanoid Whole-Body Control}, 
      author={Zixuan Chen and Mazeyu Ji and Xuxin Cheng and Xuanbin Peng and Xue Bin Peng and Xiaolong Wang},
      year={2025},
      eprint={2506.14770},
      archivePrefix={arXiv},
      primaryClass={cs.RO},
      url={https://arxiv.org/abs/2506.14770}, 
}

@misc{han2025kungfubot2,
      title={KungfuBot2: Learning Versatile Motion Skills for Humanoid Whole-Body Control}, 
      author={Jinrui Han and Weiji Xie and Jiakun Zheng and Jiyuan Shi and Weinan Zhang and Ting Xiao and Chenjia Bai},
      year={2025},
      eprint={2509.16638},
      archivePrefix={arXiv},
      primaryClass={cs.RO},
      url={https://arxiv.org/abs/2509.16638}, 
}

@conference{AMASS:ICCV:2019,
  title = {{AMASS}: Archive of Motion Capture as Surface Shapes},
  author = {Mahmood, Naureen and Ghorbani, Nima and Troje, Nikolaus F. and Pons-Moll, Gerard and Black, Michael J.},
  booktitle = {International Conference on Computer Vision},
  pages = {5442--5451},
  month = oct,
  year = {2019},
  month_numeric = {10}
}

@article{miura1984dynamic,
  title={Dynamic walk of a biped},
  author={Miura, Hirofumi and Shimoyama, Isao},
  journal={The International Journal of Robotics Research},
  volume={3},
  number={2},
  pages={60--74},
  year={1984},
  publisher={Sage Publications Sage CA: Thousand Oaks, CA}
}

@article{sreenath2011compliant,
  title={A compliant hybrid zero dynamics controller for stable, efficient and fast bipedal walking on MABEL},
  author={Sreenath, Koushil and Park, Hae-Won and Poulakakis, Ioannis and Grizzle, Jessy W},
  journal={The International Journal of Robotics Research},
  volume={30},
  number={9},
  pages={1170--1193},
  year={2011},
  publisher={SAGE Publications Sage UK: London, England}
}

@article{geyer2003positive,
  title={Positive force feedback in bouncing gaits?},
  author={Geyer, Hartmut and Seyfarth, Andre and Blickhan, Reinhard},
  journal={Proceedings of the Royal Society of London. Series B: Biological Sciences},
  volume={270},
  number={1529},
  pages={2173--2183},
  year={2003},
  publisher={The Royal Society}
}

@inproceedings{radosavovic2024humanoid,
  title={Humanoid locomotion as next token prediction},
  author={Radosavovic, Ilija and Zhang, Bike and Shi, Baifeng and Rajasegaran, Jathushan and Kamat, Sarthak and Darrell, Trevor and Sreenath, Koushil and Malik, Jitendra},
  booktitle={The Thirty-eighth Annual Conference on Neural Information Processing Systems},
  year={2024}
}

@article{radosavovic2024learning,
  title={Learning humanoid locomotion over challenging terrain},
  author={Radosavovic, Ilija and Kamat, Sarthak and Darrell, Trevor and Malik, Jitendra},
  journal={arXiv preprint arXiv:2410.03654},
  year={2024}
}

@article{radosavovic2024real,
  title={Real-world humanoid locomotion with reinforcement learning},
  author={Radosavovic, Ilija and Xiao, Tete and Zhang, Bike and Darrell, Trevor and Malik, Jitendra and Sreenath, Koushil},
  journal={Science Robotics},
  volume={9},
  number={89},
  pages={eadi9579},
  year={2024},
  publisher={American Association for the Advancement of Science}
}

@article{gu2024advancing,
  title={Advancing humanoid locomotion: Mastering challenging terrains with denoising world model learning},
  author={Gu, Xinyang and Wang, Yen-Jen and Zhu, Xiang and Shi, Chengming and Guo, Yanjiang and Liu, Yichen and Chen, Jianyu},
  journal={arXiv preprint arXiv:2408.14472},
  year={2024}
}

@article{chen2024lcp,
  title={Learning smooth humanoid locomotion through lipschitz-constrained policies},
  author={Chen, Zixuan and He, Xialin and Wang, Yen-Jen and Liao, Qiayuan and Ze, Yanjie and Li, Zhongyu and Sastry, S Shankar and Wu, Jiajun and Sreenath, Koushil and Gupta, Saurabh and others},
  journal={arXiv preprint arXiv:2410.11825},
  year={2024}
}

@article{zhang2024wococo,
  title={Wococo: Learning whole-body humanoid control with sequential contacts},
  author={Zhang, Chong and Xiao, Wenli and He, Tairan and Shi, Guanya},
  journal={arXiv preprint arXiv:2406.06005},
  year={2024}
}

@article{zhuang2024humanoidparkour,
  title={Humanoid parkour learning},
  author={Zhuang, Ziwen and Yao, Shenzhe and Zhao, Hang},
  journal={arXiv preprint arXiv:2406.10759},
  year={2024}
}

@article{xue2025hugwbc,
  title={A Unified and General Humanoid Whole-Body Controller for Fine-Grained Locomotion},
  author={Xue, Yufei and Dong, Wentao and Liu, Minghuan and Zhang, Weinan and Pang, Jiangmiao},
  journal={arXiv preprint arXiv:2502.03206},
  year={2025}
}

@article{he2025getup,
  title={Learning getting-up policies for real-world humanoid robots},
  author={He, Xialin and Dong, Runpei and Chen, Zixuan and Gupta, Saurabh},
  journal={arXiv preprint arXiv:2502.12152},
  year={2025}
}

@article{huang2025getup,
  title={Learning Humanoid Standing-up Control across Diverse Postures},
  author={Huang, Tao and Ren, Junli and Wang, Huayi and Wang, Zirui and Ben, Qingwei and Wen, Muning and Chen, Xiao and Li, Jianan and Pang, Jiangmiao},
  journal={arXiv preprint arXiv:2502.08378},
  year={2025}
}

@misc{su2025hitterhumanoidtabletennis,
      title={HITTER: A HumanoId Table TEnnis Robot via Hierarchical Planning and Learning}, 
      author={Zhi Su and Bike Zhang and Nima Rahmanian and Yuman Gao and Qiayuan Liao and Caitlin Regan and Koushil Sreenath and S. Shankar Sastry},
      year={2025},
      eprint={2508.21043},
      archivePrefix={arXiv},
      primaryClass={cs.RO},
      url={https://arxiv.org/abs/2508.21043}, 
}

@misc{ren2025humanoidgoalkeeperlearningposition,
      title={Humanoid Goalkeeper: Learning from Position Conditioned Task-Motion Constraints}, 
      author={Junli Ren and Junfeng Long and Tao Huang and Huayi Wang and Zirui Wang and Feiyu Jia and Wentao Zhang and Jingbo Wang and Ping Luo and Jiangmiao Pang},
      year={2025},
      eprint={2510.18002},
      archivePrefix={arXiv},
      primaryClass={cs.RO},
      url={https://arxiv.org/abs/2510.18002}, 
}

@inproceedings{luo2023perpetual,
  title={Perpetual humanoid control for real-time simulated avatars},
  author={Luo, Zhengyi and Cao, Jinkun and Kitani, Kris and Xu, Weipeng and others},
  booktitle={Proceedings of the IEEE/CVF International Conference on Computer Vision},
  pages={10895--10904},
  year={2023}
}

@article{fu2024humanplus,
  title={Humanplus: Humanoid shadowing and imitation from humans},
  author={Fu, Zipeng and Zhao, Qingqing and Wu, Qi and Wetzstein, Gordon and Finn, Chelsea},
  journal={arXiv preprint arXiv:2406.10454},
  year={2024}
}

@article{ppo,
  title={Proximal policy optimization algorithms},
  author={Schulman, John and Wolski, Filip and Dhariwal, Prafulla and Radford, Alec and Klimov, Oleg},
  journal={arXiv preprint arXiv:1707.06347},
  year={2017}
}

@inproceedings{ross2011reduction,
  title={A reduction of imitation learning and structured prediction to no-regret online learning},
  author={Ross, St{\'e}phane and Gordon, Geoffrey and Bagnell, Drew},
  booktitle={Proceedings of the fourteenth international conference on artificial intelligence and statistics},
  pages={627--635},
  year={2011},
  organization={JMLR Workshop and Conference Proceedings}
}

@misc{unitree-g1,
  author = {{Unitree Robotics}},
  title = {{Humanoid robot G1\_Humanoid Robot Functions\_Humanoid Robot Price | Unitree Robotics}},
  year = {2025},
  url = {https://www.unitree.com/g1/},
  note = {\url{https://www.unitree.com/g1/}}
}

@article{leon2013gram,
  title={Gram-Schmidt orthogonalization: 100 years and more},
  author={Leon, Steven J and Bj{\"o}rck, {\AA}ke and Gander, Walter},
  journal={Numerical Linear Algebra with Applications},
  volume={20},
  number={3},
  pages={492--532},
  year={2013},
  publisher={Wiley Online Library}
}

@misc{hendawy2024multitaskreinforcementlearningmixture,
      title={Multi-Task Reinforcement Learning with Mixture of Orthogonal Experts}, 
      author={Ahmed Hendawy and Jan Peters and Carlo D'Eramo},
      year={2024},
      eprint={2311.11385},
      archivePrefix={arXiv},
      primaryClass={cs.LG},
      url={https://arxiv.org/abs/2311.11385}, 
}

@article{harvey2020lafan,
    author    = {Félix G. Harvey and Mike Yurick and Derek Nowrouzezahrai and Christopher Pal},
    title     = {Robust Motion In-Betweening},
    journal = {ACM Transactions on Graphics (Proceedings of ACM SIGGRAPH)},
    publisher = {ACM},
    volume    = {39},
    number    = {4},
    year      = {2020}
}

@article{he2024h2o,
  title={Learning human-to-humanoid real-time whole-body teleoperation},
  author={He, Tairan and Luo, Zhengyi and Xiao, Wenli and Zhang, Chong and Kitani, Kris and Liu, Changliu and Shi, Guanya},
  journal={arXiv preprint arXiv:2403.04436},
  year={2024}
}

@inproceedings{Peng_2018,
   title={Sim-to-Real Transfer of Robotic Control with Dynamics Randomization},
   url={http://dx.doi.org/10.1109/ICRA.2018.8460528},
   DOI={10.1109/icra.2018.8460528},
   booktitle={2018 IEEE International Conference on Robotics and Automation (ICRA)},
   publisher={IEEE},
   author={Peng, Xue Bin and Andrychowicz, Marcin and Zaremba, Wojciech and Abbeel, Pieter},
   year={2018},
   month=may, pages={3803–3810} 
}

@article{isaac,
  title={Isaac gym: High performance gpu-based physics simulation for robot learning},
  author={Makoviychuk, Viktor and Wawrzyniak, Lukasz and Guo, Yunrong and Lu, Michelle and Storey, Kier and Macklin, Miles and Hoeller, David and Rudin, Nikita and Allshire, Arthur and Handa, Ankur and others},
  journal={arXiv preprint arXiv:2108.10470},
  year={2021}
}
